\documentclass[graybox]{svmult}

\usepackage{amsmath}
\usepackage{mathptmx}       
\usepackage{helvet}         
\usepackage{courier}        
\usepackage{type1cm}        
\usepackage{wrapfig}
\usepackage{mathptmx}       
\usepackage{helvet}         
\usepackage{courier}        
\usepackage{type1cm}        
\usepackage{graphicx}
\usepackage[table,xcdraw]{xcolor} 
 
\usepackage{makeidx}         
\usepackage{multicol}        
\usepackage[bottom]{footmisc}

\usepackage[numbers]{natbib}

\usepackage{color}
\usepackage{colortbl}
\usepackage[colorlinks]{}
\usepackage[table,xcdraw]{xcolor}
\usepackage{booktabs}
\usepackage{multirow}
\usepackage{wrapfig}

\newcommand{\probP}{\text{I\kern-0.15em P}}

\makeindex             

\begin{document}
\title*{A Transformer-based Response Evaluator for Open-Domain Spoken Conversation}
\definecolor{light-gray}{gray}{0.9}

\author{\bf Davan Harrison, Rishi Rajasekaran, and Marilyn Walker} 

\institute{Davan Harrison and Marilyn Walker \at Natural Language and Dialogue Systems Lab, UCSC, \email{\{vharriso,mawalker\}@ucsc.edu} 
\and Rishi Rajasekaran \at Amazon Alexa AI, \email{rishraja@amazon.com} \\ This work was done while Rishi Rajasekaran  was a student at UCSC.}

\maketitle

\abstract{
Many open-domain dialogue systems rely on multiple response generators, any of which can contribute a response to the dialogue in a particular context. Thus
the ability to compare potential responses and then select the best plays an important role in ensuring a dialogue system is coherent and engaging.  Dialogue coherence goes beyond simply remaining on topic -- some trivia may be on topic and engaging 
when mentioned out of the blue, but 
may not be coherent and grounded in the context of the conversation.
We carry out  experiments on response selection in  the Athena system, an Alexa Prize SocialBot that has dedicated content and multiple topic-specific response generators for a large number of topics.  First, we collect a corpus of Athena conversations with live human traffic, where potential responses from all enabled response generators are logged and subsequently annotated for response quality.
We  compare several off-the-shelf response ranking methods for open-domain dialogue  
to Athena-Heuristic, a heuristic response ranker
that was field-tested in Athena during the third Alexa Prize competition. We also compare these to   a 
transformer-based response ranker we call Athena-RR, that we train on our Athena conversations. 
Athena-RR
 uses both the conversational context and the dialogue state to rank the potential responses.  
We find that Athena-RR with a Recall@1 of 70.79\% outperforms Athena-Heuristic and all of the off-the-shelf rankers by a large margin.
We then conduct a live A/B study comparing Athena-Heuristic to Athena-RR  in a 6,358 conversations with Alexa users. We show
that Athena-RR  leads to significantly longer conversations that receive significantly higher user ratings than  the heuristic rule-based ranker.
}

\label{chap:response-evaluator}

\section{Introduction}
\label{intro-sec}

Open-domain spoken dialogue systems aim to carry out a natural conversation with a human user on any topic. While this remains an extremely challenging problem, a number of advances have been made in recent years, partly through competitions such as the Alexa Prize SocialBot Grand Challenge~\cite{khatri2018advancing,gabriel2020further}. 
However, given the diversity of user interests and utterances in conversation, even the best of these systems  can struggle to come up with good responses in  many contexts \cite{pichl2020alquist,paranjape2020neural,chen2018gunrock,fang2017sounding,curry2018alana,bowden2019entertaining}. As a result, most such systems utilize 
multiple response generators that contribute to a pool of potential responses in every dialogue  context. Thus, the ability to rank several potential responses and then select the most appropriate one plays an important role in ensuring that a dialogue system remains coherent and engaging \cite{shalyminov2018neural,hedayatnia2022systematic,ghazarian2022wrong,kim2020speech,mehri2020unsupervised,ghazarian2020predictive}. The aim of this work is to develop improved response selection methods for open-domain spoken conversation. 

The recent growth and popularity of open-domain systems has been accompanied by many new reference-free dialogue response ranking methods. Frameworks such as DialogRPT, DynaEval, USR, RUBER, and FED, {\it inter alia}, provide off-the-shelf tools  for estimating  the quality of a turn in  an ongoing conversation
\cite{gao2020dialogue, zhang2021dynaeval, yeh2021comprehensive, mehri-USR20,tao2018ruber,mehri2020unsupervised}. We test several of these off-the-shelf rankers on the task of  response ranking in Athena, a socialbot that competed in the Alexa Prize Socialbot Grand Challenge over the past few years \cite{harrison2020athena,patilathena2021}.  

\begin{figure*}[htb]
    \centering
    \fbox{\includegraphics[width=0.95\textwidth]{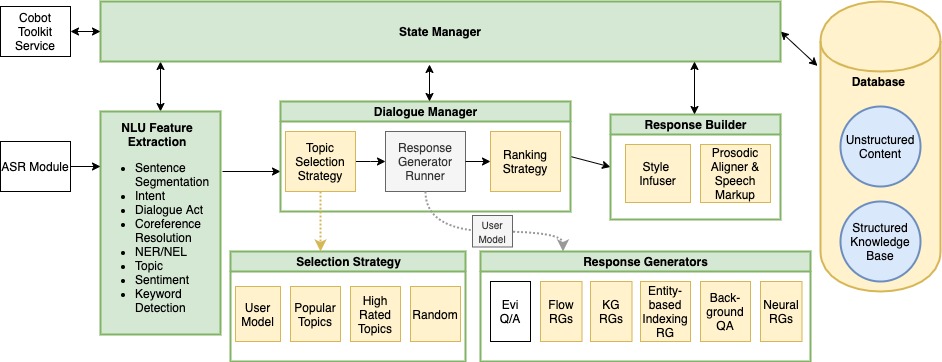}}
    \caption{Athena's architecture showing  Multiple Types of Response Generators}
    \label{fig:athena-arch}
\end{figure*}

Athena's dialogue system architecture, shown in Figure~\ref{fig:athena-arch}, follows typical socialbot design principles, where the response sub-component consists of multiple response generators (RGs) for each topic. Unlike
open-domain chatbots  based on LLMs such as GPT3 or DialoGPT  \cite{zhang2019dialogpt,radford2019language}, Athena includes topic-specific response generators that use live content from WikiData and other databases, as well as topic-specific trivia.  
For each system turn, a Dialogue Manager (DM) queries a subset of RGs for responses given the topic and the dialog context, resulting in a pool of candidate responses, from which a response is constructed and spoken to the user.  
During the third Alexa Prize competition,  we fielded Athena-Heuristic, a heuristic response ranker.
In this paper, we report experiments on training a new  transformer-based response ranker we call \textit{Athena-RR}.  We first  
create a dataset of annotated (conversation-context, response-pool) examples, from live customer traffic, by logging all the responses  that could be produced by the RGs at any point in an Athena dialogue.\footnote{We cannot release this dataset due to user privacy issues, however a checkpoint for the trained model is available at \url{github.com/Vrindiesel/athena-rr} .} We  annotate the goodness of all the  responses, 
and then use the annotated corpus to train a new response ranker. 

We  compare the performance of the DynaEval, DialogRPT, and FED 
off-the-shelf response-rankers, to Athena-Heuristic and Athena-RR on the
annotated test corpus. We also test Athena-Heuristic and Athena-RR  
in the production Athena system in an A/B study where we can measure user ratings and conversation lengths. 
We show that   Athena-RR ranker achieves an Recall@1 of 70.79\% at selecting highly rated responses, and that deploying Athena-RR to live customer traffic resulted in  an increase in mean conversation ratings  from 3.64 to 3.77 and an increase in  mean conversation length  from 15.02 to 24.77 turns. The  best off-the-shelf ranker is the DialogRPT  Depth metric with a Recall@1 of 50.56\%. We find that the DynaEval metric does not perform as well on our data as on other corpora, with a Recall@1 of only 21.34. We also find that the 8 FED metrics are very highly correlated with one another, and that their Recall@1 values range between 28.09\% and 41.57\%.

\section{Related Work}
\label{sec:background}

There has recently been an upsurge of interest in response ranking for open-domain dialogues  \cite{georgila2020predicting,deriu2021survey,yeh2021comprehensive}. In many cases, these are trained on an existing corpus, that doesn't have negative examples, so that that the ranker is trained to  select the true response as opposed to a randomly selected one \cite{lowe2017training,liu2016not}. More recent work uses quality annotations on dialogues, either from  crowd-sourced workers who created the dialogue turns, or from independent judges \cite{zhang2018personalizing,yeh2021comprehensive}, with the types of dialogues ranging
from short chit-chat conversations such as DailyDialogue \cite{li2017dailydialog}, to longer information dense conversations such as Topical Chat \cite{gopalakrishnan2019topical}. 

Our experiments test the off-the-shelf rankers DialogRPT, DynaEval, and FED. DialogRPT is trained on Reddit threads, and uses  up/down votes, number of direct responses, and length of response threads to produce different methods of ranking responses \cite{gao2020dialogue}.  DynaEval has multiple versions trained on Empathetic Dialogues, ConvAI2 and Daily Dialogue. We use the Empathetic Dialogues variant that was used by the original authors for their out-of-domain evaluation on FED\cite{zhang2022fined}.
The FED  ranker uses the pretrained transformer DialoGPT to score  responses on various dimensions of conversational quality, 
using the language model probabilities of the response eliciting a follow-on user reply such as ``That's interesting'' \cite{mehri2020unsupervised}.

There are also other response rankers specifically used in the context of the Alexa Prize.  Kim et al. uses a large corpus labelled with the output of an acoustically-based user sentiment detector to train a model that predicts the users' sentiment towards a potential  response \cite{kim2020speech}. Other work  has used specific user responses such as ``Ask me if I care'' to  label the previous turn as good or bad and then trained a ranker \cite{see2021understanding,shalyminov2018neural}. Another approach   relied on conversational data annotated at the response level for its coherence and relevance in the context \cite{yi2019towards}.  
In a similar approach to what we take here, Hedayatnia et al. trained a response ranker on an annotated set of Alexa Prize conversations with multiple responses generated by different neural response generation models (NRGs), and  then compared its performance  to  off-the-shelf models, using different methods for negative and positive sampling \cite{hedayatnia2022systematic}. In this work, there were true negative responses,
 as  in the Athena annotated corpus described in Section~\ref{sec:corpus}. However, unlike Athena, Hedayatnia's system  had only two topic-specific RGs, one for movies and another for music, and thus the response pools are quite different than Athena's. We make Athena-RR publicly available\footnote{github.com/Vrindiesel/athena-rr} in the hope that other Alexa Prize teams will test it on their systems.

\section{The Athena System}
\label{athena-sec}

Athena is an Alexa Prize socialbot that has competed in the Alexa Prize for the last two years \cite{harrison2020athena,walker2021athena}.
During the competition,
Athena carries out thousands of conversations a week with real users who access Alexa Prize systems by simply saying ``Let's Chat'' to an Alexa-enabled device.
At the end of each conversation, the user has the option to rate the conversation on a scale of 1 to 5, indicating their interest in speaking with Athena again in the future. We posited that an improved response ranker would increase Athena's average rating and lead to longer conversations.

Figure~\ref{fig:athena-arch} in Section~\ref{intro-sec}  illustrates Athena's architecture. Response ranking in Athena is especially challenging due to the large pool of RGs that can be queried in any given turn. The types of response generators (RGs) are shown in Figure~\ref{fig:athena-arch}, but there are  Flow-RGs, KG-RGs and Entity-based Indexing (trivia) RGs for each topic that Athena supports. The Dialogue Manager
tracks the discourse history and the context, as well as the topic under discussion, and uses this state information to
decide which RGs to run. 

\begin{figure*}[htb]
    \centering
    \fbox{\includegraphics[width=0.95\textwidth]{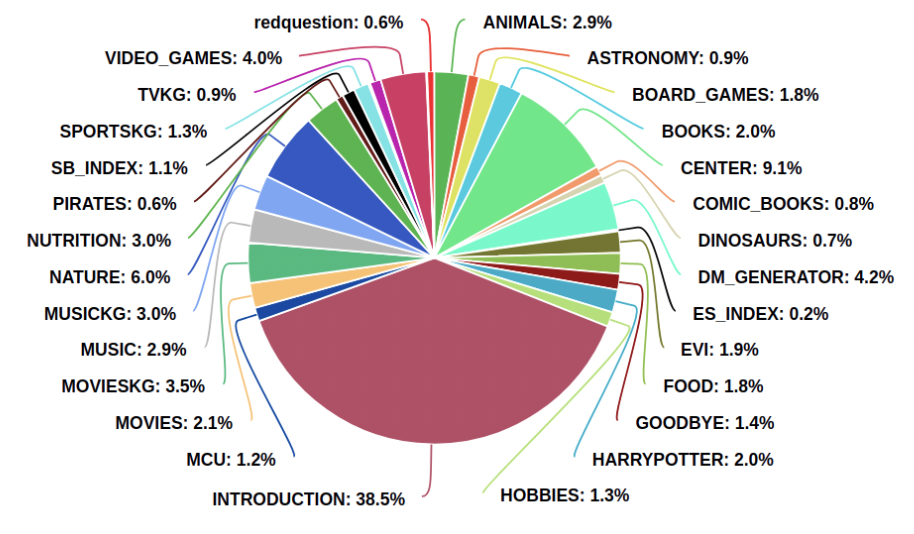}}
    \caption{Distribution of Utilization of Athena's  Response Generators over a 6 month period}
    \label{fig:rg-distribution}
\end{figure*}

Figure~\ref{fig:rg-distribution} 
shows the  utilization of Athena's RGs over a six month period. Note the large number of topics, including rare or niche topics 
such as the Marvel Comic Universe (MCU), nutrition, astronomy, pirates and dinosaurs. Our response ranker needs to work well for all of these topics. 
Figure~\ref{fig:rg-distribution} also shows  how 
topics, such as music and movies, have both a KG  and a Flow-based  RG that compete with one another
 to take the next turn. For example,  Music-KG, accesses knowledge triples about musicians and albums from the WikiData Knowledge Graph  \cite{reed2022jurassic}, while the Music RG  is a  flow-based RG  that structures the conversation  by asking the user about  their music preferences and opinions. The RG labelled as CENTER in Figure~\ref{fig:rg-distribution} 
represents the Entity Indexing RGs that provide trivia, 
with almost all  topics having a corresponding trivia RG, which could also compete for the turn.  The EVI QA engine was also called in every context to provide alternative possible responses. Neural Response Generators (NRGs) provided by Amazon were primarily run in the background in Athena due to latency issues and the difficulty with controlling their frequent topic changes \cite{rajasekaran2020,patilathena2021,chi2021neural,saha2021proto}.

\begin{figure*}[htb]
   \centering
    \fbox{\includegraphics[width=0.98\textwidth]{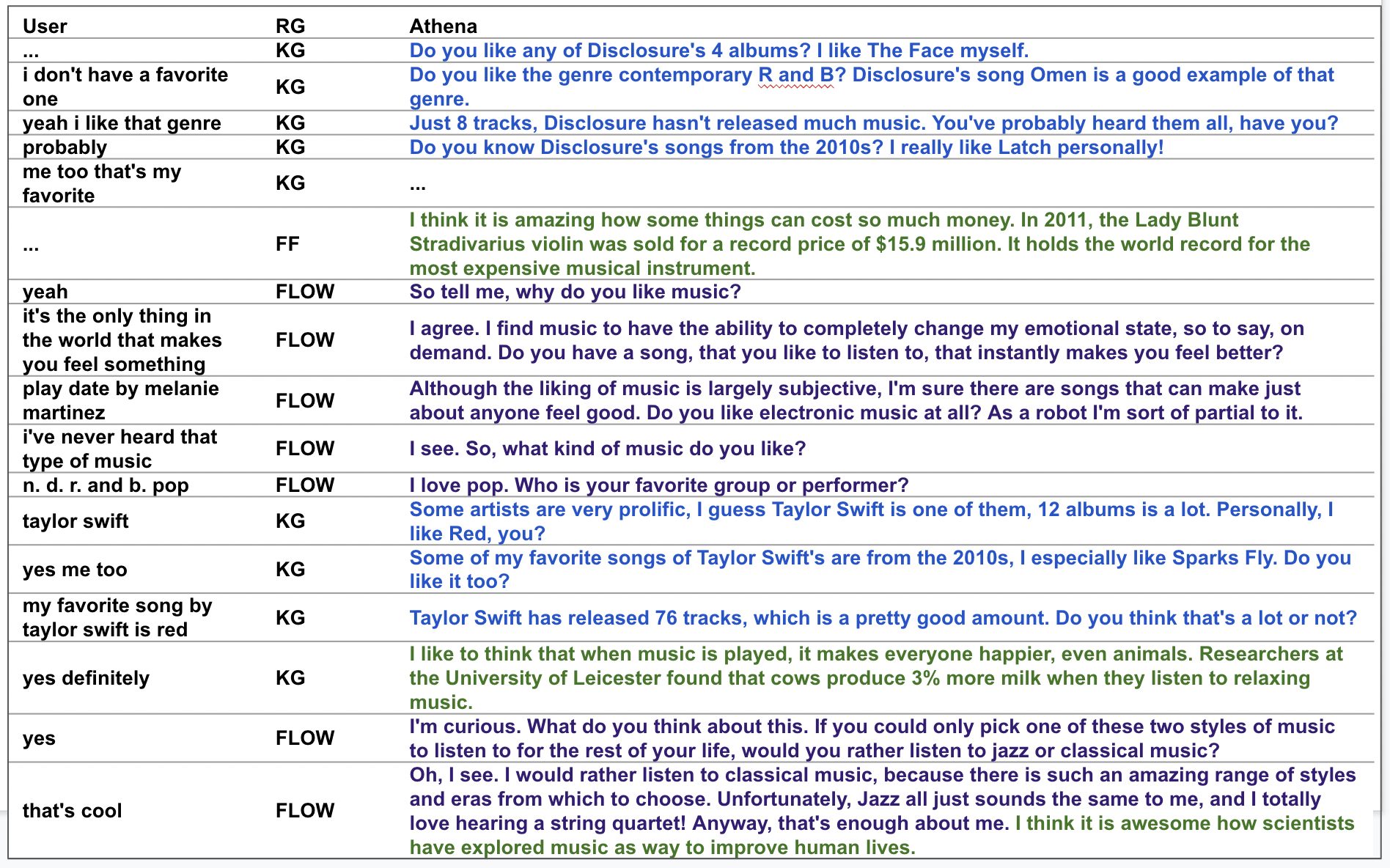}}
   \caption{Dialogue illustrating the use of Multiple Response Generators for a single topic. The RG column indicates the type of RG, also indicated through different colors for each RG type. }
   \label{fig:music-interleaving}
\end{figure*}

While this multiple-RG setup allows for flexibility in handling unplanned user behavior and results in more engaging and diverse conversations, it makes response selection given the context of the conversation a very crucial task. Figure~\ref{fig:music-interleaving} illustrates how Athena-RR, discussed in more detail below, allows the Athena system to flexibly interleave responses from the three different response generators for Music.

\section{Corpus Creation}
\label{sec:corpus}

\begin{table}[ht]
    \centering
    \resizebox{\textwidth}{!}{%
    \begin{tabular}{l|p{3.0in}|c} 
    \toprule
        {\cellcolor[gray]{0.9} \textbf{Speaker}} & {\cellcolor[gray]{0.9} \textbf{Dialogue Context}}  & {\cellcolor[gray]{0.9} \bf Response Generator} \\ \toprule
        ATHENA & Yeah. A lot of people have cool hobbies, but what about me? What kind of hobby do you think I would like? & Intro RG\\ \hline
        USER & what's the name for music &\\ \hline
        ATHENA &  I love talking about music!  Is there a musician or band that you like? & Music-KG\\ \hline
        USER & no \\ \bottomrule
        { \cellcolor[gray]{0.9}  \textbf{Rank}} & { \cellcolor[gray]{0.9}  \textbf{Response Candidates}} & { \cellcolor[gray]{0.9}  \textbf{Response Generator}}\\ \toprule
        D &  I recently read this fact about Britney Spears. Spears had gotten her start in fame through being on The Mickey Mouse Club with other artists like Christina Aguilera.  & Center-Music \\ \hline
        A & no problem! it can be hard to think of one.   i like talking about the band  Fleetwood Mac.  Do you like Fleetwood Mac? & Music-KG \\ \hline
        C &  I recently read an interesting fact about the world's largest piano. The record for the largest piano ever made is held by Adrian Mann, a piano tuner from New Zealand. It took him four years to build the piano, and this grandiose instrument weighs 1.4 tonnes and is 5.7 meters long.  & Center-Music \\ \hline
        A & So tell me, why do you like music? & Music-Flow\\ \hline
        C &  I just read this fascinating fact about the history of Drums. Did you know that drums are one of the oldest musical instruments in the world? & Center-Music \\ \hline
        C &  Do you know the size of the world's smallest violin is? Well,  I heard in Guangzhou, China, a 66-year-old senior violinist Chen Lianzhi made a 1-centimeter-long violin which sets the new world record for the world's smallest violin.  & Center-Music \\ \hline
        ~ & None of the above. \\ \hline
    \end{tabular}
    }
        \caption{Response pool annotation task example for Music. An ``A" indicates an excellent response, while an``B" indicates a response that could be used, and a ``C" indicates a response that might be okay in another context, while a ``D" indicates a bad response. }
    
    \label{fig:rr-response-pool-music-anno-example}
\end{table}

We use Athena's live traffic with real human users to create our annotated corpus.
We take a random sample of 1,600 response pool instances with their associated conversation histories. The 1,600 instances are comprised of 100 random samples from each of 16 internal topics used by Athena, as illustrated in Figure~\ref{fig:rg-distribution}. The responses are selected such that we get 20 instances from conversations belonging to each 1-5 rating value, whenever possible.

\begin{table}[ht]
    \centering
\resizebox{\textwidth}{!}{%
    \begin{tabular}{l|p{3.0in}|c} 
   {\cellcolor[gray]{0.9} \textbf{Speaker}} & {\cellcolor[gray]{0.9} \textbf{Dialogue Context}}  & {\cellcolor[gray]{0.9} \bf Response Generator} \\ 
    
	ATHENA	&	Nice choice. And for me? It's hard for me to pick a specific scene, but I certainly liked Eli Wallach in this movie. What did you think of her portrayal of Tuco?		& Movies-Flow	\\ \hline																			
	USER	&	i'm not familiar with that actress	&	\\ \hline																				
	ATHENA	&	In order to capture legal conversations, the Director of Marriage story would meet with actual divorce attorneys. &	Center-Movies\\ \hline																					
	USER	&	i don't know anything about that &	\\ \bottomrule
    {\cellcolor[gray]{0.9}      \textbf{Rank}} & {\cellcolor[gray]{0.9}  \textbf{Response Candidate Text}} & {\cellcolor[gray]{0.9} \bf Response Generator}  \\ \toprule																					
	A	&	Seen any good movies recently?	& Movies-KG		\\ \hline																			
	A	&	I was wondering what you think about this. Would you rather star in a commercial blockbuster or an independent film that wins prestigious awards?	& Movies-Flow		\\ \hline																			
	C	&	Show business seems like it would be a really confusing business to navigate. Keanu Reeves was actually tricked into appearing in the film when a friend forged his signature on the contract.&	Center-Movies \\ \hline																					
	C	&	I already though Charlize Theron was really interesting, and then I heard this. When she was a baby, Charlize Theron had jaundice, and the doctors put her on antibiotics. Her baby teeth rotted because of the medicine, so she did not have a normal set of teeth until she was 11 years old and her adult teeth grew in.&	Center-Movies \\ \hline																					
	B	&	The director would explain the concept of the divorce at the center of the story then ask what kind of things these lawyers would target in the characters.	& Center-Movies	\\ \hline																				
	A	&	Who is your favorite actor or actress?	& Movies-KG		\\ \hline																			
	C	&	Is there a director that you like?  Is there a movie that you want to talk about? &	Movies-KG	\\ \hline																				
	C	&	I was a little disappointed that The Two Popes had some historical inaccuracies. One thing The Two Popes got right though is that Pope Benedict liked Orange Fanta, he drank at least 4 cans of the soda every day!&	Center-Movies \\ \hline																					
		&	None of the above.	& None-candidate	\\ \hline
			\end{tabular}
   }
        \caption{Response pool annotation task example for Movies. An ``A" indicates an excellent response, while an``B" indicates a response that could be used, and a ``C" indicates a response that might be okay in another context, while a ``D" indicates a bad response.  }

    \label{fig:rr-response-pool-movies-anno-example}
\end{table}

Athena particularly struggles with selecting good responses to questions, so we also bias the sampling to include 40/100 instances where the user's most recent utterance is a question. This is not always possible because question dialogue acts occur less frequently within some topics compared to others. So we collect an additional random sample of 494 instances of users asking questions, without regards to topic. This amounts to a total of 2,094 response pool instances. When collecting response pools, we limit the sampling to instances where the response pool contains more than one candidate response. An example response pool for a Music conversation is shown in Table~\ref{fig:rr-response-pool-music-anno-example} and one for Movies is shown in Table~\ref{fig:rr-response-pool-movies-anno-example}. 

We collect annotations for the corpus via  a response selection task where human judges rate response candidates. In each annotation instance, the judges are presented with 4 turns of conversation history ending with the user's most recent utterance, with a pool of response candidates listed below. The data is in spreadsheet format. Annotators are asked to select a response that they would say next by filling in the cell next to the selected response candidate.  The annotations for the two  response pool examples are shown in the LHS of Table~\ref{fig:rr-response-pool-music-anno-example} and of Table~\ref{fig:rr-response-pool-movies-anno-example}.  Annotators are allowed to select multiple response candidates if multiple responses would be appropriate to say next. Notice that in both Table~\ref{fig:rr-response-pool-music-anno-example} and  Table~\ref{fig:rr-response-pool-movies-anno-example} multiple responses are indicated with an ``A'' indicating they are equally good. Notice also that the response pools show how different RGs for the same topic compete with one another to take the turn. The Center RGs provide trivia, and they are part of the response pool in general to guarantee that Athena always has something to say, but even though they are ``on topic'' they are often not as good as responses produced by the KG response generators or the scripted Flow response generators.
Annotators are always given a  "None of the above" option for situations where the response pool does not contain any good candidates. The annotations are performed by the authors and recruited lab-mates. Each response pool received one annotation therefore we did not measure annotator agreement. 

\begin{table}[hbt]
\centering
\resizebox{\textwidth}{!}{%
\begin{tabular}{lr|lr|lr}
\toprule
\multicolumn{2}{c|}{\cellcolor[gray]{0.9} Corpus Descriptive Statistics} 
& \multicolumn{2}{c}{\cellcolor[gray]{0.9} Un-normalized Corpus}
& \multicolumn{2}{c}{\cellcolor[gray]{0.9} Normalized Corpus}
\\
\toprule
\# topics & 16  & \# annotated responses  & 7,915 & \# annotated responses & 6,113 \\

\# instances per topic  & 100    
& mean response pool size  & 4.43 
& mean response pool size  & 3.42 \\

\# topical instances  & 1,600   
& median response pool size & 4.0 & median response pool size  & 3.0
\\

\# extra question instances & 494  & & &  &
\\

Total \# of instances  & 2,094   & & & & \\
\bottomrule
\end{tabular}
}
\caption{Corpus statistics for annotated Athena response pool dataset.}
\label{table:rr-athena-corpus-stats}
\end{table}

Overall statistics on the annotated corpus of Athena response pools can be seen in Table~\ref{table:rr-athena-corpus-stats}
and Table~\ref{table:rr-athena-corpus-stats2}. Table~\ref{table:rr-athena-corpus-stats} summarizes how the response pool covers all of Athena's topics, as well as a general pool of questions. The mean and median response pool sizes are artificially reduced to make it easier for the annotators. In general, in any dialogue context, there are many more than 3 or 4 possible responses as illustrated by the response pool annotation examples in Figure~\ref{fig:rr-response-pool-music-anno-example} and Figure \ref{fig:rr-response-pool-movies-anno-example}. This is because there are many Center-RG potential responses as well as multiple responses from the knowledge graph RGs.

\begin{table}[t]
\centering
\begin{tabular}{l|r|r}
\toprule
{\cellcolor[gray]{0.9} Data} & {\cellcolor[gray]{0.9} Topical instances} & {\cellcolor[gray]{0.9}   Question instances}\\ \hline
\# initial response pools &  1,600 (\%100.0)  &  \\
\# pools with only negative responses  & 306  (\%19.1)  & - \\
\# pools with positive responses & 1,294  (\%80.8)  & 494  \\
\bottomrule
\# included in training set & 1,294 &  494 \\
\bottomrule

\end{tabular}
\caption{Response Pools positive and negative instances}
\label{table:rr-athena-corpus-stats2}
\end{table}

Table~\ref{table:rr-athena-corpus-stats2} breaks the corpus statistics down further, to show how many positive and  negative examples there are in each category of topical instances and question instances. Despite the many different RGs, 19.1\% of the response pools are judged to have no good response, perhaps illustrating how it can be difficult to fully predict how users will behave. This observation is also supported  by other work on Alexa Prize, where approximately  33\% of  the response pools had  no good responses \cite{hedayatnia2022systematic}.

\section{Ranking Responses: Experiment Overview}
\label{sec:ranking-models-intro}

Our primary goal is to evaluate how well different response ranking methods, when given the conversation context, are able to select the best response out of a pool of responses. We train our own Athena specific response ranker Athena-RR, and test three recently proposed models for
domain independent, reference free response ranking, namely  DialogRPT, DynaEval and FED \cite{zhang2022fined,mehri2020unsupervised,gao2020dialogue}.
We compare these to Athena-Heuristic, a response ranker that was fielded in the third Alexa Prize competition. We evaluate response ranking methods in two ways: (1) accuracy against our test corpus; and  (2) by deployment into the production Athena system as part of an A/B study on user ratings and conversation lengths. 

\subsection{Response Ranking Models}
\label{response-models-sec}

\vspace{.1in}
\noindent{\bf Athena's Heuristic Response Ranker.}
Athena's heuristic response ranker is based on a topic-agnostic approach to dialogue management based on theories of discourse coherence~\cite{GroszCentering,prasad2010exploiting} to dynamically structure a dialogue. 
The heuristic ranker applies rules on the current dialogue state to try to remain on topic until the user requests a topic change or the topic has been exhausted.
Topic-specific RGs can provide additional signals to the DM along with their response such as an indication that they \textit{must continue}, \textit{can continue}, or \textit{have ended} their conversation.
Preference is first given to the RG that had control on the last turn, in order to increase local coherence and promote conversation flow.
Then preference would be given to another RG on the same topic, for example when a FlowRG for music was exhausted, the KG-RG for music
and the Center RG for music would be equally preferred.

\vspace{.1in}
\noindent{\bf DialogRPT Response Ranker.}
\label{sec:dialogRPT-model}
DialogRPT\footnote{\url{https://github.com/golsun/DialogRPT}}\cite{gao2020dialogue}
is a transformer-based dialogue response ranking model trained by predicting human feedback (i.e., upvotes or replies) of responses on a large-scale social media platform. The model was trained on 100's of millions of Reddit
conversation threads. DialogRPT is trained with a contrastive ranking loss where the model learns to  perform pairwise classification, maximizing the likelihood of the positive class instance over that of the negative class instance. 

Multiple versions of DialogRPT were trained using different types of human feedback. We use three variants of DialogRPT whose checkpoints were made publicly available by the original authors. The DialogRPT-updown was trained using post upvote information, DialogRPT-width and DialogRPT-depth were trained using conversation thread width and depth information, respectively. We use the implementation and model available through Huggingface Tranformers.\footnote{\url{https://github.com/huggingface/transformers}}

\vspace{.1in}
\noindent{\bf DynaEval Ranker.}
DynaEval \cite{zhang2022fined} leverages a Graph Neural Network to model
the dialog-level interactions between a user and a
system. This allows it  to distinguish well-formed dialogues from carefully constructed negative samples that allow it to better capture dialogue level coherence. Previous work that tested DynaEval on
DailyDialog, the USR subset of Topical Chat, ConvAI2 and Empathetic Dialogues showed
that it performs considerably better than other metrics across these different datasets \cite{yeh2021comprehensive}. We evaluate DynaEval for turn-level response ranking and selection by selecting the candidate response with the highest dialog-level coherence score.  

\vspace{.1in}
\noindent{\bf The FED Ranker.}
\label{sec:fed-ranker}
FED  is a method proposed to perform reference-free evaluation of 18 fine-grained dialogue qualities using a large pretrained model \cite{mehri2020unsupervised}. These dialogue qualities are further divided into 8 \textit{turn-level} and 10 \textit{conversation-level} qualities. The turn-level qualities measure whether a specific turn of a conversation was \textit{Interesting, Engaging, Specific, Relevant, Correct, Semantically Appropriate, Understandable,} and \textit{Fluent}. We select FED as one of our baselines for turn-level evaluation in Athena because the Meena and Mitsuku datasets that it showed good results for  are  similar to Athena conversations \cite{adiwardana2020towards}. 

\begin{table}
\begin{footnotesize}
\begin{tabular}{p{.8in} p{.5in} p{3.0in}}
\toprule
 Metric    &  Type & Probe \\ \midrule
  Interesting  &
         Positive   &  Wow that is really interesting. / That's really interesting! / Cool! That sounds super interesting.   \\
        & Negative   &  That's not very interesting. /  That's really boring. /  That was a really boring response.   \\ \midrule
     Engaging   &
         Positive  &   Wow! That's really cool! /  Tell me more! /
                         I'm really interested in learning more about this.   \\
       &  Negative   &   Let's change the topic. /  I don't really care.  That's pretty boring. / I want to talk about something else.  
                        \\ \midrule
    
     Relevant   &
         Positive   & \; ---  \\
        & Negative   &  That's not even related to what I said. /  Don't change the topic! / Why are you changing the topic?  
     \\ \midrule
     Correct   &
         Positive  &  \; ---  \\
        & Negative   &  You're not understanding me! / I am so confused right now! /
                         I don't understand what you're saying.  
     \\ \midrule
     Semantically   &
       Positive   &  That makes sense. /  You have a good point.  , \\
        Appropriate    & negative  &   That makes no sense!  
     \\ \midrule
     Fluent   &
         Positive   &  That makes sense! /  You have a good point.   \\
        & Negative   &  Is that real English? /  I'm so confused right now! /  That makes no sense!  
 \\ \bottomrule

\end{tabular}
\caption{\label{fed-probes} Sample User response probes for FED turn level metrics}

\end{footnotesize}
\end{table}

The turn-level metrics are computed by determining the relative likelihood of a system response in context being followed by a set of predefined positive and negative probe responses for a given dialoguequality. The probe responses
are shown in Table~\ref{fed-probes}.
For each conversation in the normalized corpus, the turn-level FED metric scores given the previous context is computed for each candidate response using a \textit{DialoGPT-large} model. We exclude dialogues that do not have any context turns, since the initial turns are greeting turns and do not involve response selection. This yields a final corpus of 5927 candidate responses in 1743 dialogues.

FED is meant to provide general metrics based on LLMs that are not a reflection of a particular training set.  Previous work suggests that the correlations between FED metrics and  
human annotations can be as high as .21 at the turn level and as high as .443 at the dialogue level \cite{mehri2020unsupervised}. 
Our hypothesis was that the different turn-level FED measures strongly overlap in terms of what they are trying to capture.

\begin{wrapfigure}{r}{2.5in}
\vspace{-.35in}
    \centering
    \includegraphics[width=2.5in]{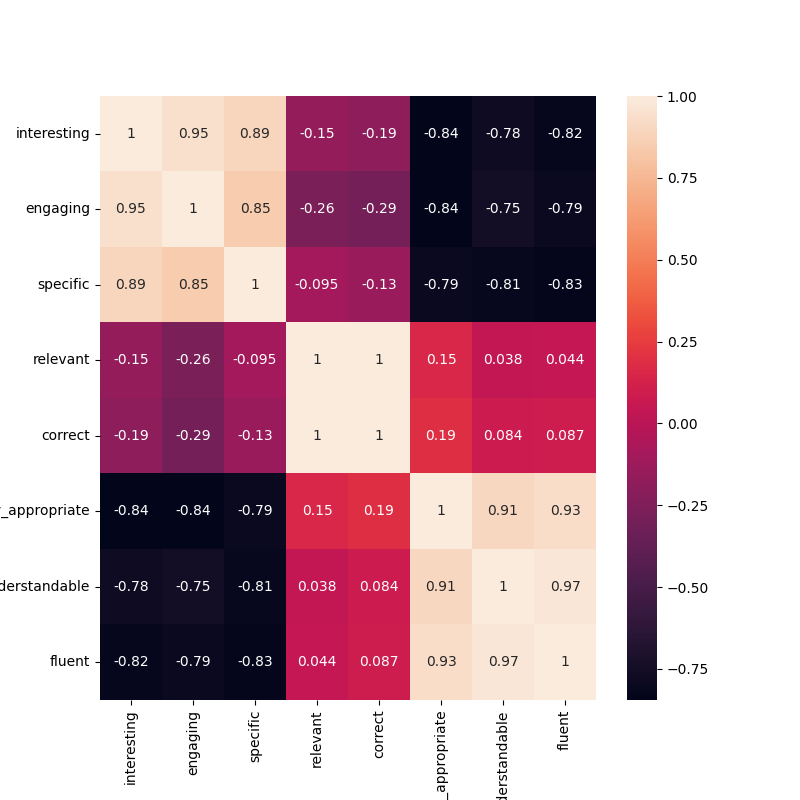}
    \caption{FED metric Correlations on Athena corpus (DialoGPT-large)}
    \label{fig:rr-fed_correlation_matrix_large}
\vspace{-.2in}
\end{wrapfigure}

To explore this, we calculated the the Pearson correlation for the individual metric scores
for all the responses from the conversation corpus. The correlation heatmap is shown in Figure~\ref{fig:rr-fed_correlation_matrix_large}. The heatmap clearly shows (light cream blocks) that sets of metrics are highly correlated with one another. In particular, the following groups of metrics exhibit a high degree of correlation with each other --- (i) \textit{interesting}, \textit{engaging}, \textit{specific} (ii) \textit{relevant}, \textit{correct}, (iii) \textit{semantically appropriate}, \textit{understandable}, \textit{fluent}. The heatmap also shows that  the \textit{interesting}, \textit{engaging}, and \textit{specific} metrics have a strong negative correlation with the \textit{semantically appropriate}, \textit{understandable}, \textit{fluent} metrics.

We posit two causes for this result --- one being the conversations that DialoGPT is trained on are very different from Athena conversations and the other being many of the chosen reference sentences used to compute the FED scores are very similar across the correlated metrics. Note that the reference sentences shown in Table~\ref{fed-probes} for \textit{semantically appropriate, understandable}, and \textit{fluent} are based on the same positive prompts while using very similar or overlapping examples of negative prompts. Also  \textit{relevant, correct} both contain only negative prompts while \textit{interesting, engaging, specific} use positive prompts that can be construed as being similar. As a result,  the metrics in each correlated group appear to measure the same attribute, which is in line with related work that report similar correlations in the human annotation scores of the conversation-level qualities in the FED dataset\cite{zhang2022fined}.

We also explored whether  there's a significant difference between the FED metric scores for the population of human preferred and dispreferred responses in the normalized Athena corpus. We did a two-sided t-test  for the FED metrics between the samples of human preferred and dispreferred utterances. We found that \textit{Interesting, Engaging, Specific} scores for the human preferred responses were actually significantly lower than those of dispreferred responses; in other words they go the opposite direction of what they are supposed to measure. The two samples for \textit{Relevant, Correct, Semantically Appropriate, Fluent} showed no statistically significant difference. The only metric that worked as predicted was  \textit{Understandable}, where the scores for human preferred responses were significantly higher than those of the dispreferred responses. 
Overall, the FED metrics demonstrate weak results on Athena conversations for response ranking and may not be a suitable measure for dialogue quality evaluations. 

\vspace{.1in}
\noindent{\bf Athena-RR: A Response Ranking transformer.} 
\label{sec:Athena-RR-model}

Much of the recent research using pre-trained transformer models use an end-to-end dialogue learning approach. Rather than aiming to  construct an end-to-end model, we want to enhance the dialogue system that we already have. Therefore, we will not be able to ``copy" the  problem setup used in other work \cite{Gopalakrishnan2019,Dinan_Roller_Shuster_Fan_Auli_Weston_2019,Wolf_Sanh_Chaumond_Delangue_2019}.
We also want Athena's response ranker to take advantage of all the information contained in Athena's internal dialogue state representation.  We develop a method of incorporating the additional information into the model by extending the input sequence dialogue tokens using special token representations of Athena's internal state.

We train a BERT based model, Athena-RR, using our annotated corpus of Athena's response pools. Athena-RR is trained by optimizing a next-utterance classification loss, similar to that used in the next-sentence prediction task used in BERT\cite{Devlin2019BERTPO}. The model performs binary classification on each response candidate to predict whether or not a potential utterance occurs next for a given conversation context. Our end goal is to have a model capable of picking responses from a response pool rather than just performing binary classification on responses.

Our model is pre-trained on the Topical Chats corpus using random sampling for False next-utterances. We take the model checkpoint with the highest validation accuracy on the Topical Chats corpus and fine-tune the model on Athena conversation data. For the sake of evaluation on this intermediate binary classification task, we use a random training/evaluation partition of the dataset using a $90:10$ ratio. 

To smooth the distribution of positive and negative class instances in the data, the annotated corpus is down sampled to smooth the distributions of response candidate selections. Some response candidates were presented to annotators a disproportionate number of times, resulting in some utterances having been not selected most of the times they were in a response pool. This is due to the internal working of Athena, and the way that the response pools were collected. In cases where an example occurs more than 8 times as a negative example, we cap its frequency using the cap value $N_{\text{cap}}$, computed as:
$ N_{\text{cap}} = 8 + 2 \cdot \text{log}(\bar{C}) $
where $\bar{C}$ is the number of times annotators had the opportunity to select the candidate but didn't.

\vspace{.1in}

\noindent
\textbf{Input Representation.}
Model inputs are the conversation context and a candidate response.
Conversation context includes four turns of history represented as sequence of tokens with a special \texttt{[SEP]} token between turns. The  context also includes special tokens that represent Athena's internal dialogue state, such as the current topic and the previous topic. 
We also add special tokens to the front of the response candidate token sequence that denote the RG it came from.

\begin{figure}[htb]
    \centering
    \includegraphics[width=\textwidth]{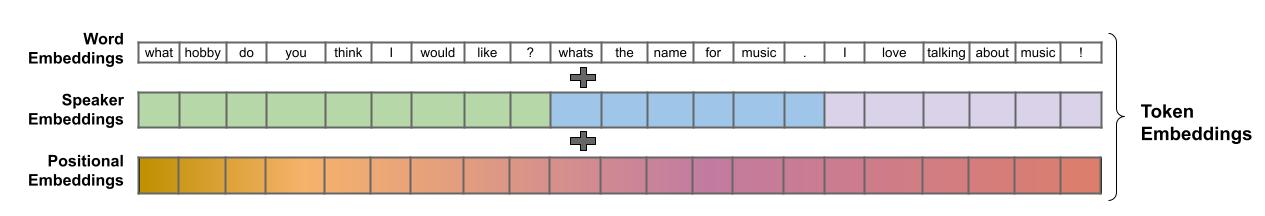}
    \caption{Speaker aware token embeddings.}
    \label{fig:speaker-aware-embeddings}
\end{figure}
\noindent
\textbf{Speaker-Aware turn embeddings.}
Athena-RR is implemented using the HuggingFace Transformers\cite{Wolf2019HuggingFacesTS} 
implementation of the BERT transformer in our experiments.
Similar to \citep{Wolf_Sanh_Chaumond_Delangue_2019}, we adapt the model by adding special embeddings that indicate the speaker and the dialogue state information. Figure~\ref{fig:speaker-aware-embeddings} shows the speaker aware embedding token and embedding function. 

\subsection{Evaluation Methods and Setup}

\noindent{\bf Corpus Evaluation.}
We evaluate all five ranking models, DialogRPT, DynaEval, FED,  Athena-Heuristic and Athena-RR in terms of recall@1 on a held-out test set of 89 (context, response pool) instances. 
To measure the performance of the off-the-shelf models,  we test whether the response with the highest model score is a human preferred response in each conversation.

\vspace{+.1in}
\noindent{\bf Evaluation: Live User Study.}
Because Athena is a real system that interacts with human users, we were also able to 
incorporate our response ranking model into Athena's dialogue manager and deploy it in an A/B study with the  \textit{A} version using a pre-existing heuristic response ranker, and the \textit{B} version using Athena-RR. 
We trained a new version of the model using our entire annotation corpus including our test set. During training, the model learns to perform binary classification on a given response candidate: say it next or don't say it next. We incorporate the model into Athena by scoring each candidate response separately for a given response pool, and then returning the response with the highest log-likelihood. Note that Athena-RR does not do topic-selection, this is driven by a model of user topic preferences and the dialogue context. Athena-RR is also not used when the response pool contains only one response candidate.

\section{Results}
\label{sec:results}

\begin{wraptable}{r}{2.0in}
\vspace{-0.3in}
\centering
\begin{small}
\begin{tabular}{lr}
\toprule
Model       &  Recall@1 \\ \midrule
DialogRPT-updown & 29.21\% \\
DialogRPT-width & 46.07\% \\
DialogRPT-depth & 47.19\% \\
DialogRPT-human-vs-rand & 42.70\% \\
DialogRPT-human-vs-machine & 38.20\% \\
\midrule
DynaEval &  21.34\% \\ \midrule  
FED-Interesting  & 28.09\% \\ 
FED-Engaging & 30.34\% \\
FED-Specific & 26.97\% \\
FED-Relevant & 39.33\% \\
FED-Correct & 41.57\% \\
FED-Semantically Appropriate & 41.57\% \\
FED-Understandable & 40.45\% \\
FED-Fluent & 39.33\%  \\ \midrule
Athena-Heuristic & 56.18\% \\ \midrule
Athena-RR & 70.79\%  \\ 
\bottomrule
\end{tabular}
\end{small}
\caption{Performance  of the various response rankers on the test set}
\label{table:ranking-autoeval}
\end{wraptable}

Table~\ref{table:ranking-autoeval} shows that DynaEval and the DialogRPT-updown metric have the poorest performance with accuracies of only 21.34\% and 29.21\% respectively.
The DialogRPT width and depth metrics perform much better with the depth metric achieving an accuracy of 47.19\%. DialogRPT is better than the best FED metric. Previous work suggests that the DialogRPT human-vs-random ranker performs the best \cite{hedayatnia2022systematic}, though the human-based metrics are not better than DialogRPT-width and DialogRPT-depth on the Athena corpus. The FED metrics \textit{Interesting, Engaging, Specific} all have accuracies at 30\% or worse.  \textit{Relevant, Correct, Semantically Appropriate, Understandable}, and \textit{Fluent} have the highest accuracies, with best accuracy at 41.57\%, but the differences between them are not statistically significant (p $=$ .58).  The Athena-Heuristic ranker is better than DialogRPT-depth and better than  FED Semantically Appropriate (p $=$ 0.037), which is the highest accuracy FED metric. The relatively poorer performance of the FED metrics in comparison to the Athena-Heuristic ranker that simply tries to stay on topic indicates that FED does not have any notion of topic relevance. The Athena-RR ranker, trained specifically for this task, performs the best, as might be expected. The accuracy of 70.79\% is much better than that achieved by the previous heuristic  Athena response ranker.

\begin{wraptable}{r}{2.0in}
\vspace{-.2in}
\begin{small}
\centering
\begin{tabular}{lcccc}
           &     & Median  & Mean    & Mean     \\
Version    & N    & \# turns & \# turns   &  rating \\ \toprule
Heur.  & 3,502 & 10  & 15.02  & 3.64  \\
Trans. & 2,856 & 19 & 24.77 & 3.77  \\ \hline
           &      &  & \textit{p}\textless 0.01 & \textit{p}\textless 0.01 \\ \bottomrule
\end{tabular}
\caption{A/B study results: heuristic response ranker versus AthenaRR.}
\label{table:rr-ab-study-results}
\end{small}
\end{wraptable}
\vspace{.1in}
\noindent{\bf Human Evaluation A/B Study}. We then deployed Athena-RR and Athena-Heuristic simultaneously in a live A/B study. Because the metrics used in the Alexa Prize are  the average  user rating and length (number of turns) for conversations, we evaluate those measures in the A/B study. Athena-RR is not used until the fourth system turn so we only consider conversations that meet this length requirement. We also note that there are a number of possible situations in a conversation where response ranking is not used \cite{whang2021response}. For instance, if the user says "Let's talk about something else", the flow-of-control returns to the DM to select a new topic. Also, Athena-RR is not used when the system needs to perform a functional speech act, e.g., user asks the system to repeat itself, or Athena needs to deflect a user seeking adult content. 

Table~\ref{table:rr-ab-study-results} shows the A/B study results for conversations where Athena-RR was used on one or more turn, compared to conversations where Athena-Heuristic  was used on every turn. The results show that Athena-RR outperforms the Athena-Heuristic model for both conversation length and ratings, with a significant (p $<$ .01) and  large increase in both measures.

\section{Discussion and Conclusion}
\label{sec:discussion}
\label{sec:conclusion}

This paper describes a response ranker for the Athena SocialBot system, a competitor in the Alexa Prize. Response ranking is challenging in Athena because it has the capacity to provide many different on-topic responses in a particular dialogue context: the response ranker needs to discriminate between different on-topic responses. We developed  a corpus of Athena conversations with real human users by logging all possible responses that Athena could produce for many different dialogue contexts, and then sampled these so as to represent all of Athena's topics, as well as to have a good distribution of contexts where users ask questions.

We then explore  different  models for response ranking and selection. We train a new  conversational response ranking model, Athena-RR,  based on the speaker-aware BERT architecture and compare its performance to the reference-free dialogue evaluation metrics  DialogRPT, DynaEval and FED, as well as to Athena-Heuristic, the rule-based response ranker fielded in Alexa Prize 3. We find that DialogRPT-depth provides the highest performance of the off-the-shelf metrics with a Recall@1 of 47.19\%. We show that DynEval performs much worse on Athena than on other corpora \cite{yeh2021comprehensive}. We also demonstrate that the turn-level metrics within  FED exhibit strong correlations with each other, as also found in previous work \cite{zhang2022fined}, and that  FED performs significantly worse than DialogRPT-depth, with the best turn-level FED metrics  achieving a Recall@1 of 41.57\%. Finally, we show that Athena-RR significantly outperforms  Athena-Heuristic  in an offline evaluation with a Recall@1 of 70.79\% as compared to 56.18\%, and that it is also significantly better than any of the off-the-shelf metrics.

One surprising result  was that Athena-Heuristic, which simply tries to stay on topic,  significantly outperformed all the off-the-shelf metrics (p $=$ 0).  One would imagine that DialogRPT, which is trained on  thousands of Reddit conversations, would have internalized a notion of topical coherence. Perhaps DialogRPT  performs more poorly than expected because Reddit conversations are  very different in style than Athena: the conversations are between humans, topical coherence may not be an issue, and the number of  human responses to an utterance may reflect their individual opinions rather than utterance quality.

Finally we conduct an A/B study of Athena-RR and Athena-Heuristic in our live deployed system with over 6000 conversations with end-users. Our results show that Athena-RR   led to significantly better conversation ratings (3.77 compared to 3.64) and longer conversations (24.77 turns as compared to 15.02 turns) when compared to those chosen by a heuristic-based response ranker.

\bibliographystyle{plainnat}

\end{document}